%% file: main.tex
\pdfoutput=1
\documentclass[11pt]{article}

\usepackage[preprint]{acl}

\usepackage{times}
\usepackage{latexsym}
\usepackage{amssymb}
\usepackage{subfigure}

\usepackage[T1]{fontenc}

\usepackage[utf8]{inputenc}

\usepackage{microtype}

\usepackage{inconsolata}

\usepackage{graphicx}
\usepackage{amsmath}
\usepackage{bm}
\usepackage{algorithm}
\usepackage{algorithmic}
\usepackage{enumitem}

\usepackage{url}

\DeclareMathOperator*{\argmin}{argmin}
    
\newcommand{\method}{FlexiGPT}

\title{FlexiGPT: Pruning and Extending Large Language Models\\ with Low-Rank Weight Sharing}

\author{
  \textbf{James Seale Smith\textsuperscript{1} \quad Chi-Heng Lin\textsuperscript{1} \quad Shikhar Tuli\textsuperscript{1} \quad Haris Jeelani\textsuperscript{1}} \\
   \textbf{Shangqian Gao\textsuperscript{2} \quad Yilin Shen\textsuperscript{1} \quad Hongxia Jin\textsuperscript{1} \quad Yen-Chang Hsu\textsuperscript{1}} \\
  \textsuperscript{1}Samsung Research America, \textsuperscript{2}Florida State University \\
}

\begin{document}
\maketitle

\input{sections/0_abstract}

\input{sections/1_intro.tex}
\input{sections/2_related.tex}

\input{sections/3_method}

\input{sections/4_experiments-prune}
\input{sections/5_experiments-extend}

\input{sections/6_conclusion.tex}

\bibliography{references}

\input{figures/method_prune_svd-rank-analysis_b}
\newpage

\section*{Appendix}
\setcounter{figure}{0}
\setcounter{table}{0}
\renewcommand{\thetable}{\Alph{table}}
\renewcommand{\thefigure}{\Alph{figure}}
\renewcommand\thesection{\Alph{section}}
\appendix
\input{sections/7_appendix}
\end{document}

%% file: sections/0_abstract.tex
\begin{abstract}
The rapid proliferation of large language models (LLMs) in natural language processing (NLP) has created a critical need for techniques that enable efficient deployment on memory-constrained devices without compromising performance. We present a method to prune LLMs that selectively prunes model blocks based on an importance score and replaces them with a low-parameter replacement strategy. Specifically, we propose a principled metric to replace each pruned block using a weight-sharing mechanism that leverages unpruned counterparts from the model and block-specific low-rank adapters. Furthermore, we facilitate the learning of these replacement blocks with output feature normalization and an adapter initialization scheme built on low-rank SVD reconstructions. Empirical evaluations demonstrate substantial performance gains over existing methods, achieving state-of-the-art performance on 5/6 benchmarks for a compression rate of $30\%$ and 6/6 benchmarks for a compression rate of $40\%$. We also demonstrate that our approach can extend smaller models, boosting performance on 6/6 benchmarks using only $\approx$0.3\% tokens of extended training with minimal additional parameter costs.
\end{abstract}

%% file: sections/1_intro.tex
\section{Introduction}
\label{sec:intro}

The widespread adoption of LLMs has revolutionized NLP applications, driving significant advancements in areas such as virtual assistants, automated customer support, and real-time language translation~\cite{minaee2024large,naveed2023comprehensive}. However, deploying these models on memory-constrained devices, such as smartphones and edge devices, remains a formidable challenge due to their substantial parameter sizes and computational demands~\cite{hadi2023large,raiaan2024review}. This paper addresses this challenge by presenting a novel approach that targets \emph{parameter efficiency} to make LLMs more suitable for on-device applications with minimal performance compromises.

Parameter efficiency is particularly critical as it directly impacts the feasibility of deploying LLMs on devices with limited memory and storage resources. Recent model pruning techniques, such as SliceGPT~\cite{ashkboos2024slicegpt}, LLM Surgeon~\cite{van2023llm}, LLM-Pruner~\cite{ma2023llm}, LaCo~\cite{yang2024laco}, and ShortGPT~\cite{men2024shortgpt}, reduce the number of parameters but often result in significant performance degradation with minimal recovery after pruning. This gap in existing techniques underscores the need for an end-to-end pruning method that not only reduces the model size but also facilitates performance recovery. \emph{In this work, we propose to recover performance by utilizing existing weights within the model.}

\input{figures/teaser_fig}

Specifically, we introduce a comprehensive pruning strategy combined with an innovative weight sharing technique and Low-Rank Adapters (LoRA)~\cite{hu2021lora}, facilitating efficient parameter usage while preserving the model's performance. We begin by pruning model blocks based on ShortGPT's Block Influence (BI) score~\cite{men2024shortgpt}. To replace the pruned blocks, we introduce a low-parameter weight-sharing mechanism that leverages existing block modules within the model and incorporates block-specific LoRA parameters, ensuring the selected replacement blocks have high similarity to the pruned blocks while maintaining block diversity. Furthermore, we introduce a novel method to initialize the LoRA adapters in weight-sharing blocks, setting them to be the low-rank difference between the pruned block and the weight-shared replacement block. This initialization minimizes initial disruptions and facilitates smoother model adaptation. Finally, we incorporate \emph{output} feature normalization for pruned blocks to ensure a smooth transition and adaptation, allowing the model to gradually learn and stabilize its performance over time. 

Empirical evaluations of our method, which we refer to as \emph{\method}, demonstrate substantial performance gains over existing methods. Specifically, we achieve state-of-the-art performance on 5/6 benchmarks for a compression rate of $30\%$ and 6/6 benchmarks for a compression rate of $40\%$ for the popular LLaMA-2 7B model~\cite{touvron2023llama}. As visualized in Figure~\ref{fig:teaser}, our proposed technique not only effectively prunes large models for on-device deployment but \emph{also extends smaller models}, improving their performance at minimal additional parameter costs. Specifically, our method shows that a 22-layer TinyLLaMA~\cite{zhang2024tinyllama} model can be extended with repeated blocks, boosting performance on 6/6 benchmarks using only $\approx$0.3\% tokens of extended training with minimal additional parameter costs. \emph{In summary, we make the following contributions}:
\begin{enumerate}
[topsep=0pt,itemsep=-1ex,partopsep=1ex,parsep=1ex,leftmargin=*,labelindent=0pt]
    \item We develop a weight-sharing technique using adapters and low-rank SVD reconstructions to replace pruned blocks effectively.
    \item We apply output normalization to maintain stability and enable gradual learning post-pruning.
    \item We propose a method for extending smaller models by repeating layers with unique adapters and normalization parameters.
    \item We achieve significant empirical performance gains, achieving state-of-the-art performance on several benchmarks for a variety of models.
\end{enumerate}

%% file: figures/teaser_fig.tex
\begin{figure}[t]
    \centering
    \includegraphics[page=2,width=0.47\textwidth,trim={1.3cm 5.3cm 15.8cm 0cm},clip]{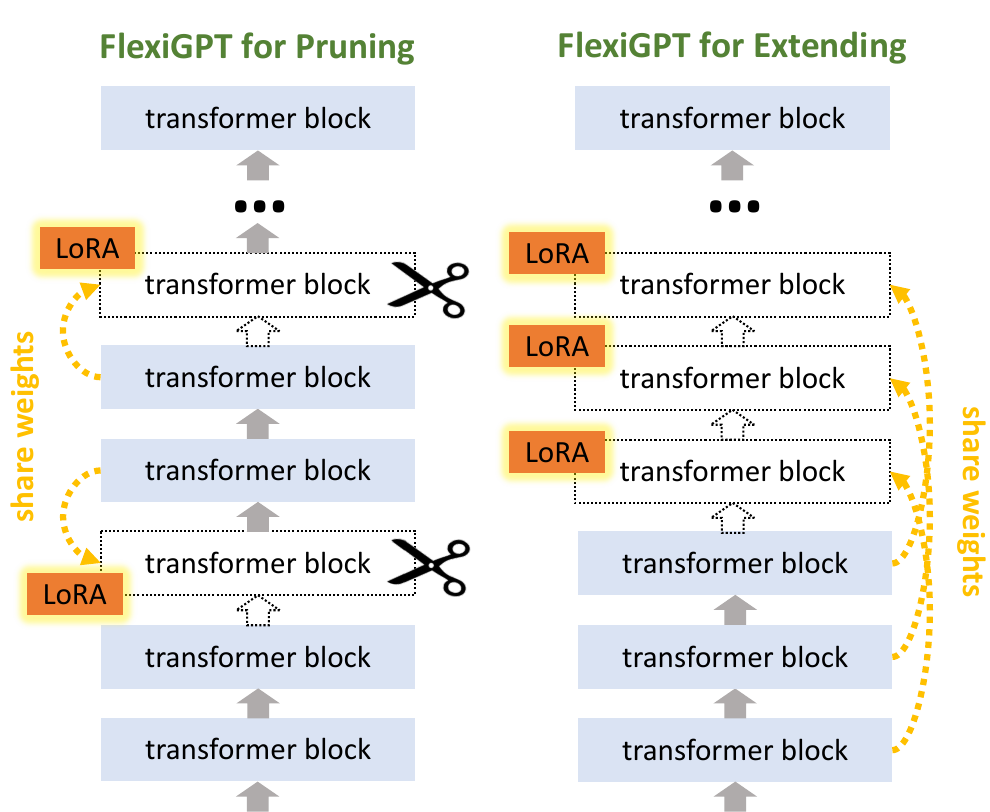}
    \caption{FlexiGPT is used for two settings: (1) pruning a model to reduce parameters with minimal performance cost \emph{or} (2) extending a model to increase performance with minimal parameter cost. \textbf{Left:} For pruning models (setting 1), we prune entire blocks and replace them using weight sharing and learned adapters. \textbf{Right:} For extending models (setting 2), we repeat block patterns in the model using weight sharing and learned adapters.
    } 
    \label{fig:teaser}

\end{figure}

%% file: sections/2_related.tex
\section{Background and Related Work}
\label{sec:rl}

\noindent \textbf{Pruning} - Pruning selectively removes less important parameters, reducing model size and computational complexity while maintaining performance. Its greatest benefit lies in optimizing LLMs for deployment in resource-constrained environments such as mobile devices, facilitating faster inference. Several works have been proposed to reduce the size of LLMs by pruning model structures. LLM-Pruner~\cite{ma2023llm} removes unimportant coupled structures and the importance is calculated from Taylor expansions. SliceGPT~\cite{ashkboos2024slicegpt} applies orthogonal projections to the feature maps and then it performs pruning in the projected space. LLM Surgeon~\cite{van2023llm} periodically updates model weights and structures, resulting in a higher cost compared to other methods. Besides reducing the width of the model, ShortGPT~\cite{men2024shortgpt} is proposed to remove blocks by using Block Influence scores, and LaCo~\cite{yang2024laco} is proposed to collapse layers. Existing pruning methods primarily focus on removing redundant model weights, often neglecting the loss of model capacity. Our approach addresses this limitation by sharing weights from the pruned model to restore its capacity.

\noindent \textbf{PEFT} - Parameter-Efficient Fine-Tuning (PEFT) methods aim to mitigate the extensive computational and memory demands of fine-tuning large models by focusing on a smaller subset of parameters. One prominent category of PEFT methods is \emph{adapters}, which involves adding trainable modules to the existing frozen layers of the model~\cite{he2021towards,houlsby2019parameter}. Another significant category is \emph{prompt} methods, which augment the initial input sequence with additional trainable vectors known as prompts. This technique focuses on fine-tuning these added tokens rather than the entire model, as demonstrated in works such as~\citet{lester2021power, liu2021p}.

Recently, LoRA~\cite{hu2021lora} has emerged as the most efficient and highest-performing PEFT approach. LoRA introduces the use of low-rank matrices to adjust model weights efficiently, merging with pre-trained weights before inference to maintain the model's operational speed. Building on this, DoRA~\cite{liu2024dora} decomposes pre-trained weights into magnitude and direction components for fine-tuning, focusing on fine-tuning the directional components using LoRA. \emph{Our research extends these low-rank PEFT methods} by incorporating LoRA and normalization for efficient weight sharing. Similar to DoRA, our method involves a normalization stage; however, we normalize in the \emph{feature-space} instead of the \emph{weight-space}.

\noindent \textbf{Weight-Sharing} - 
For GPT LLMs, \citet{dehghani2018universal} proposed to share all the layers with a dynamic halting mechanism to improve accuracy on the downstream tasks. However, it requires the number of parameters of the base layer (unshared) to match the number of parameters of all layers of vanilla transformers \cite{vaswani2017attention}. Subformer \cite{reid2021subformer} applies a sandwich-like method of parameter sharing where only middle layers are shared but it does not use any adapter. ~\cite{takase2021lessons} developed efficient cyclic sharing patterns to increase the accuracy, however their sharing patterns are mainly based on ablation studies. 
MobileLLM~\cite{liu2024mobilellm} proposed sub-billion parameter architectures for mobile devices and adopted immediate block-wise weight sharing for further accuracy improvement. However, they do not use any pruning and adapters. \citet{cao2024head} introduced matching functions to develop head-wise shareable attention in a principled fashion. Although they use pretrained weights for faster convergence, their matching functions are only applied to share weights among multiple heads of the same layer.

\noindent \textbf{SVD} - The Singular Value Decomposition (SVD) of a matrix \( \mathbf{W} \in \mathbb{R}^{m \times n} \)  is, \(\mathbf{W} = \mathbf{U} \mathbf{\Sigma}\mathbf{V}^T \), where \( \mathbf{U} \in \mathbb{R}^{m \times m} \) and \( V \in \mathbb{R}^{n \times n} \) are orthogonal matrices, and \( \mathbf{\Sigma} \in \mathbb{R}^{m \times n} \) is a diagonal matrix of singular values. SVD is widely used to obtain a low-rank representation of \(\mathbf{W}\) by selecting the \( k \) most significant singular values and their corresponding singular vectors, where \( k < \min(m,n) \). Hence, the low-rank representation of \(\mathbf{W}\) is given as: \( \mathbf{W}_k = \mathbf{U}_k \mathbf{\Sigma}_k \mathbf{V}_k^T\), where \( \mathbf{U}_k \in \mathbb{R}^{m \times k} \), \( \mathbf{\Sigma}_k \in \mathbb{R}^{k \times k} \), and \( \mathbf{V}_k \in \mathbb{R}^{n \times k} \), have lower dimensions than \(\mathbf{U,\Sigma, V}\), respectively. SVD has been applied for model compression~\cite{denton2014exploiting, hsu2022language} and is closely related to LoRA methods for reducing fine-tuning overhead.

%% file: sections/3_method.tex
\section{\method}
\label{sec:method}
\input{figures/method_prune_svd-rank-analysis}

In this section, we detail our approach, FlexiGPT (Figure~\ref{fig:method}), which  prunes and extends LLMs using LoRA adapters, weight sharing techniques, and output feature normalization. Our method focuses on achieving parameter efficiency while minimizing performance degredation, particularly for memory-constrained devices. 

Our method is based on the transformer architecture~\cite{vaswani2017attention}, which consists of Multi-Head Self-Attention (MHSA) and Multi-Layer Perceptron (MLP) layers. However, our approach is not constrained to this architecture. In general, we refer to blocks (MHSA+MLP), layers (MHSA or MLP), and weights (denoted as $W$), as our method affects the weights in a uniform manner.

\subsection{Pruning Strategy}
\label{sec:method:prune}

Our pruning strategy aims to identify and remove blocks that minimally impact the model's performance. To achieve this, we leverage the ShortGPT~\cite{men2024shortgpt} Block Influence (BI) score, which has been shown to effectively measure the importance of each block. The Block Influence (BI) score~\cite{men2024shortgpt} \(\text{BI}_i\) for a block \(i\) is defined as follows:

\begin{align}
\text{BI}_i = 1 - \mathbb{E}_{X,t} \frac{X_{i,t}^TX_{i+1,t}}{||X_{i,t}||_2||X_{i+1,t}||_2},
\label{eq:prune-score}
\end{align}
\looseness=-1
where $X_{i,t}$ denotes the $t^{th}$ row of $X_i$, and $X_i$ represents the hidden states matrix at block $i$, with dimensions $T \times d$, where $T$ is the sequence length and $d$ is the hidden dimension. This score captures the extent to which each block transforms its input, with higher scores indicating more significant changes. We calculate the BI score for each block in our model using the validation MiniPile~\cite{kaddour2023minipile} subset of the Pile dataset~\cite{gao2020pile}, and prune the blocks with the lowest BI scores. 

We tried other criteria to select blocks for pruning that considered a block's replaceability by another block in the model. However, we found that the BI score results in higher performance on downstream tasks. Our intuition is that the BI score prunes blocks deeper along the model's depth in a sequence, leaving much of the model intact and in the same order, which may explain how it retains strong downstream performance.

\subsection{Selection of Weight Sharing Bases}
\label{sec:method:index}

To replace pruned blocks, we aim to find similar unpruned blocks in the model which, when paired with adapters, can recover much of the performance lost after pruning. We aim to select each pruned block's weight sharing `base' by identifying \emph{similar} unpruned weights. However, a naïve approach such as the Frobenius norm in the weight space often results in suboptimal selections. Specifically, we find that \emph{all blocks `choose' a single block}, whereas intuition suggests a diverse selection of base blocks would work better\footnote{We empirically demonstrate this in the experiments section in the first row of Table~\ref{tab:ablations_ppl}}.

Instead, we employ a selection metric based on \emph{low-rank SVD reconstructions} to achieve a more effective and intuitive solution. Utilizing low-rank approximations, namely $\mathbf{\hat{W}}_i$ and $\mathbf{\hat{W}}_j$, instead of directly using $\mathbf{W}_i$ and $\mathbf{W}_j$ helps avoid the pitfall where all pruned blocks are replaced by a single block. We believe high-rank elimination is beneficial because low-rank approximations capture the most significant components of the weights, thereby simplifying the process of identifying suitable replacements by eliminating high-rank `noise'. Our method reveals that blocks nearest to the pruned blocks in the model tend to have the lowest scores, indicating higher similarity.

The distance metric \(d(\mathbf{W}_i, \mathbf{W}_j)\) for selecting the replacement block is defined as:

\begin{align}
& d(\mathbf{W}_i, \mathbf{W}_j) =\left\|\mathbf{\hat{W}}_i - \left(\mathbf{\hat{W}}_j + \Delta_{i-j}\right)\right\|_F
\label{eq:ws-score}
\end{align}
where:
\begin{itemize}
    \item \(\mathbf{\hat{W}}_i\) and \(\mathbf{\hat{W}}_j\) are the low-rank SVD reconstructions of \(\mathbf{W}_i\) and \(\mathbf{W}_j\), respectively, using the first \(r\) ranks.
    \item \(\Delta_{i-j} \triangleq (\mathbf{U}_{i-j} \mathbf{\Sigma}_{i-j})[1:r] (\mathbf{V}_{i-j}[1:r])^T\) is the rank-$r$ approximation of the difference \(\mathbf{\hat{W}}_i - \mathbf{\hat{W}}_j\). We used a rank of $r=256$.
    \item \(\|\cdot\|_F\) denotes the Frobenius norm.
\end{itemize}

These low-rank approximations \(\mathbf{\hat{W}}_i\) and \(\mathbf{\hat{W}}_j\) are obtained via:
\begin{equation}
\mathbf{\hat{W}}_i = ( \mathbf{U}_i \mathbf{\Sigma}_i)[1:r]  (\mathbf{V}_i[1:r])^T
\end{equation}
\begin{equation}
\mathbf{\hat{W}}_j = ( \mathbf{U}_j \mathbf{\Sigma}_j)[1:r]  (\mathbf{V}_j[1:r])^T
\end{equation}

Finally, for each pruned block \(i\), we select its base for weight sharing as the candidate block \(j\) with the minimum score in~\eqref{eq:ws-score}:
\begin{equation}
j = \argmin_{j' \neq i} d(\mathbf{W}i, \mathbf{W}{j'})
\label{eq:select-base-score}
\end{equation}

This approach is highly intuitive, as proximal blocks are naturally more alike. In Figure~\ref{fig:prune-svd-rank-analysis-a}, the proposed metric with high-rank pruning, Eq. \eqref{eq:ws-score}, shows that blocks closer in the model score lower (better), confirming our intuition that proximate blocks have similar functions. Figures~\ref{fig:prune-svd-rank-analysis-b} and \ref{fig:prune-svd-rank-analysis-c}, which respectively ablate high-rank pruning and use simple Frobenius norm, lack this clear trend, and furthermore we found that they result in significantly weaker models. An alternative version of this Figure is available in the Appendix, where the x-axis is candidate block index $j$ instead of block index distance ($i-j$).

\subsection{Output Normalization}
\label{sec:method:norm}
\input{figures/method_fig}

We apply layer normalization~\cite{ba2016layer} to the output of each MHSA and MLP layer in the weight-sharing layers, specifically to the previously pruned blocks. This normalization is applied across the hidden state dimension and is initialized to a small value set by a hyperparameter, allowing the model to gradually learn and adjust the output magnitudes over time. The normalized output \(\mathbf{h}_{norm}\) is defined as:

\begin{equation}
\mathbf{h}_{norm} = \frac{\mathbf{h} - \mu(\mathbf{h})}{\sigma(\mathbf{h})} \times \gamma
\label{eq:out-norm}
\end{equation}
where:
\begin{itemize}
    \item \(\mathbf{h}\) is the output of the layer before normalization.
    \item \(\mu(\mathbf{h})\) and \(\sigma(\mathbf{h})\) are the mean and standard deviation of \(\mathbf{h}\), respectively.
    \item \(\gamma\) is a learnable scaling weight of the same dimension as the model hidden state size.
\end{itemize}

This approach is akin to initializing the \(B\) matrix in LoRA~\cite{hu2021lora} such that \( \Delta W=BA\) is zero at the beginning of training. This similarity arises because both methods aim to minimize initial disruptions to the model and allow gradual learning. In LoRA, initializing \( \Delta W=BA\) to zero helps avoid high initial loss, ensuring smoother training. Similarly, by initializing the hidden states of weight-shared blocks to small values, we avoid significant jumps in PPL at the start of training. As shown in Table~\ref{tab:ablations_ppl}, this approach is crucial for maintaining low PPL post-pruning and ensuring stable model performance during fine-tuning.

\subsection{Adapters and Initialization}
\label{sec:method:adapters}

We employ LoRA to facilitate weight sharing for the pruned blocks, providing a parameter-efficient mechanism to adjust the weights of the replaced blocks. The LoRA adapters consist of two low-rank matrices, \(A\) and \(B\), inserted into the linear transformations of the shared weights in the model, effectively increasing the expressive capacity of these blocks despite the weight-sharing constraint\footnote{We also `unlock` the shared weight during training.}. The weights of the adapters are initialized using the SVD between the pruned block and its replacement block, as described in the selection of weight-sharing bases. Specifically, we decompose the difference between the pruned block \(\mathbf{W}_i\) and the replacement block \(\mathbf{W}_j\) into low-rank matrices:

\begin{equation}
\mathbf{W}_i - \mathbf{W}_j = \mathbf{U}_{i-j} \mathbf{\Sigma}_{i-j} \mathbf{V}_{i-j}^T
\end{equation}

The adapter matrices \(A\) and \(B\) are then initialized as:

\begin{equation}
A = (\mathbf{U}_{i-j} \mathbf{\Sigma}_{i-j})[1:r],  B = (\mathbf{V}_{i-j}[1:r])^T
\end{equation}
where:
\begin{itemize}
    \item \((\mathbf{U}_{i-j} \mathbf{\Sigma}_{i-j})[1:r]\) is the product of the left singular vectors and the diagonal matrix of singular values, indexed to take the first \(r\) columns.
    \item \((\mathbf{V}_{i-j}[1:r])^T\) is the transposed matrix containing the first \(r\) columns of the right singular vectors.
\end{itemize}

Our method requires a small amount of post-pruning fine-tuning to fully recover performance, which is discussed in Section~\ref{sec:exp_prune}. However, we generally observe that the post-prune PPL is indicative of which method will finish with a lower PPL. In Table~\ref{tab:ablations_ppl}, we see the effect of output normalization and LoRA initialization on post-prune PPL. While the SVD initialization is of smaller yet significant importance to our method, the output normalization, initialized to a small value to minimize initial disruptions and allow gradual learning, is crucially important. This is evident from the drastic increase in post-prune PPL when output normalization is ablated. The combination of SVD initialization and carefully tuned output normalization ensures that our method maintains low perplexity and stable performance during the fine-tuning phase.

\subsection{Model Extension}
\label{sec:method:extension}

In addition to pruning, FlexiGPT can also be used to extend smaller models, such as a 22-layer TinyLLaMA~\cite{zhang2024tinyllama}. In this second setting, we repeat blocks in a sequence determined by hyperparameter indexes that denote the start and end of the repetition. For instance, we might repeat layers indexed 3 through 18. Each repeated block has unique LoRA adapters and normalization parameters, and we apply output normalization to repeated blocks after the first repetition. We explore two repetition patterns: (i) \textbf{block}: each block is repeated a specified number of times, and (ii) \textbf{sequential}: the entire sequence of blocks is repeated in a specified manner. This method allows for efficient extension of smaller models, improving their performance while introducing minimal parameter overhead.

%% file: figures/method_prune_svd-rank-analysis.tex
\begin{figure*}[t]
    \centering
    \subfigure[Eq. \eqref{eq:ws-score} \emph{with} high-rank pruning]{
        \includegraphics[width=0.31\textwidth]{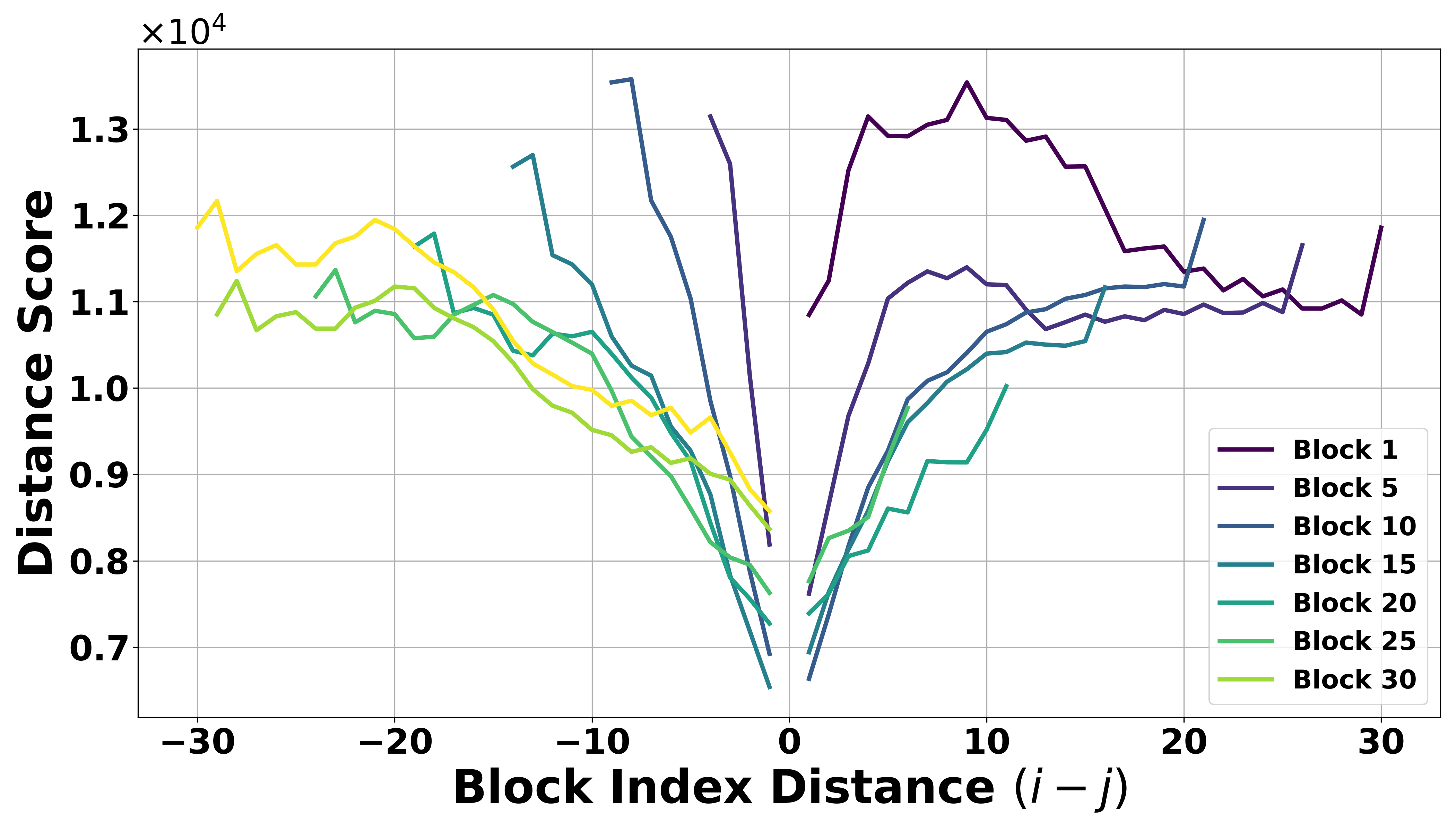}
        \label{fig:prune-svd-rank-analysis-a}
    }
    \hfill
    \subfigure[Eq. \eqref{eq:ws-score} \emph{without} high-rank pruning]{
        \includegraphics[width=0.31\textwidth]{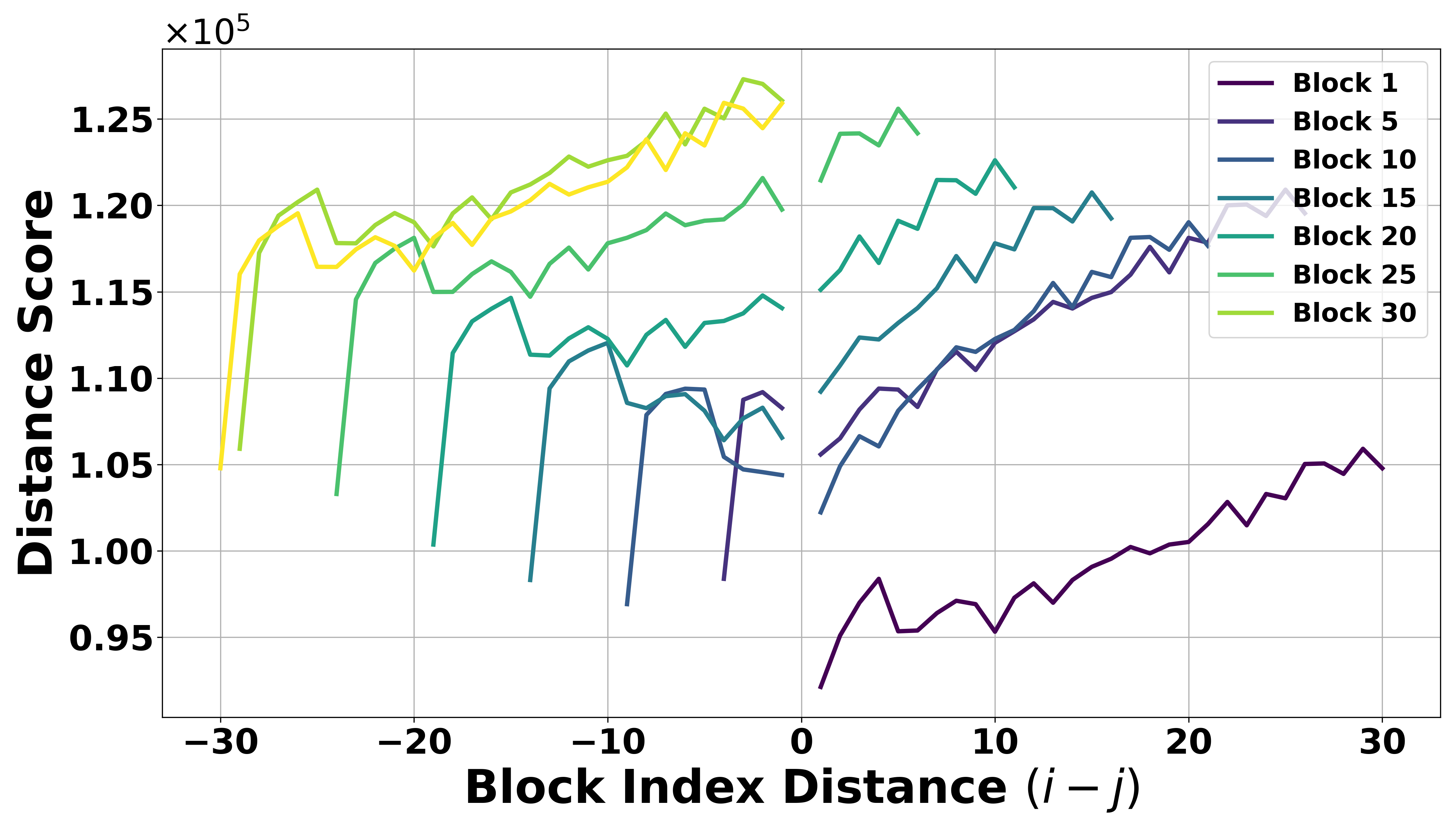}
        \label{fig:prune-svd-rank-analysis-b}
    }
    \hfill
    \subfigure[Frobenius norm of $\mathbf{W}_i - \mathbf{W}_j$]{
        \includegraphics[width=0.31\textwidth]{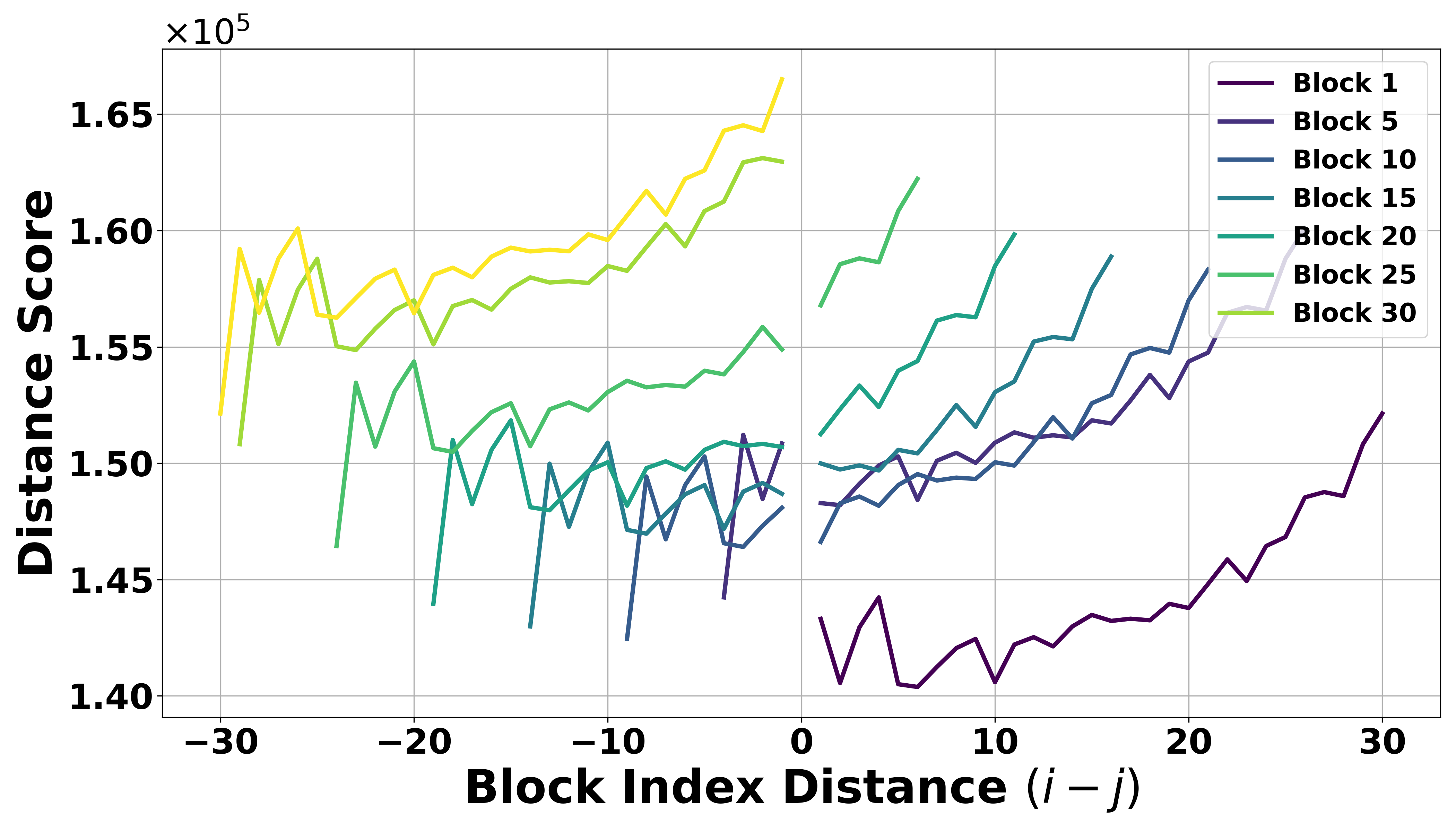}
        \label{fig:prune-svd-rank-analysis-c}
    }
    \caption{Comparison of block distance score versus block index distance ($i-j$) for different metrics. (a) Using the proposed metric in Eq. \eqref{eq:ws-score} with high-rank pruning, showing that closer blocks score lower (better), matching our intuition that weights close in the model have similar function. (b) Ablation of high-rank pruning, where there is no clear trend except that blocks closer to 0 are lower and those closer to 31 are higher. (c) Simple Frobenius norm, showing a similar lack of clear trend as in (b). \emph{We found that using the score in (a) as the weight-sharing selection metric results in a much higher performing model compared to using the scores in (b) and (c).}}
    \label{fig:prune-svd-rank-analysis}
\end{figure*}

%% file: figures/method_fig.tex
\begin{figure*}[t]
    \centering
    \includegraphics[page=1,width=\textwidth,trim={1.2cm 3.5cm .2cm 0cm},clip]{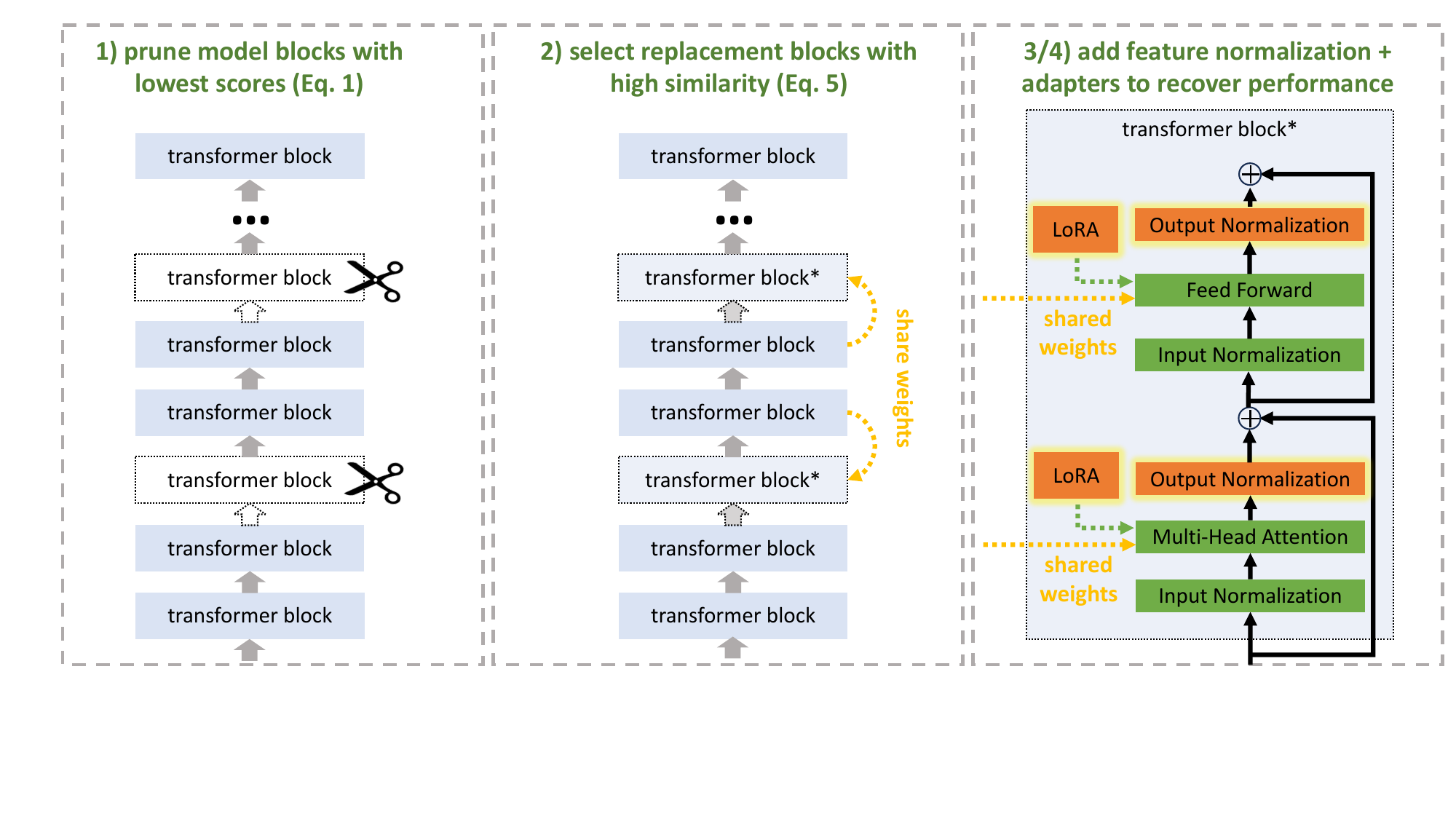}
     \caption{Overview of the FlexiGPT pruning process. \textbf{Left:} We prune model blocks with the lowest scores based on \eqref{eq:prune-score}. \textbf{Center:} We select replacement blocks with high similarity using \eqref{eq:select-base-score}. \textbf{Right:} We add feature normalization and learn adapters to recover performance.
    }
    \label{fig:method}
\end{figure*}

%% file: sections/4_experiments-prune.tex
\section{Model Compression with FlexiGPT}
\label{sec:exp_prune}
\input{tables/main-prune}
\input{tables/main-extend}

\subsection{Setup}
\noindent \textbf{Models} - We evaluated our method using LLaMA-2 7B~\cite{touvron2023llama}, OPT 1.3B and 6.7B~\cite{zhang2022opt}, and LLaMA-3 8B~\cite{llama3modelcard}, focusing on these due to their widespread adoption by the community.

\noindent \textbf{Frameworks and resources} - Our implementations were done using PyTorch, leveraging FSDP and FP-16 mixed training for efficiency. Experiments were conducted on 4 NVIDIA A100 80GB GPUs, and we utilized the Hugging Face Transformers library for model handling and training. Detailed configurations and additional resources are provided in the Appendix.

\noindent \textbf{Datasets and Benchmarks} - We used 1B tokens from the SlimPajama~\cite{cerebras2023slimpajama} pre-training dataset for post-prune recovery.  For zero-shot performance evaluations, we use the ARC-e, ARC-c \cite{clark2018think}, PIQA \cite{bisk2020piqa}, WinoGrande \cite{sakaguchi2021winogrande}, and HellaSwag \cite{zellers2019hellaswag} zero-shot benchmarks, utilizing the LM Evaluation Harness~\cite{gao2021framework}. For perplexity performance evaluations, we use the validation MiniPile~\cite{kaddour2023minipile} subset of the Pile dataset~\cite{gao2020pile}\footnote{We avoid the SlimPajama validation set to avoid giving an unfair advantage to methods trained on the this dataset.}. %

\noindent \textbf{Baselines} - We compared our method against several baselines, including LLM Surgeon \cite{van2023llm}, SliceGPT \cite{ashkboos2024slicegpt}, ShortGPT \cite{men2024shortgpt}, and ShortGPT + LoRA (an improved version of ShortGPT for a fair comparison with our method). LLM Surgeon and SliceGPT are presented for additional context for experiment results which overlapped with our setting (we use the original results presented in their papers), whereas we implement ShortGPT from scratch for a direct comparison in our setting. LLM-Pruner~\cite{ma2023llm} and LaCo~\cite{yang2024laco} are not included in our tables as ShortGPT has been found to outperform both methods.

\subsection{Results}

\noindent \textbf{Main Results} - Table~\ref{tab:main_results_prune} summarizes the perplexity (PPL) and zero-shot task performance of various pruning methods on the Llama-2 7B model. Our method, FlexiGPT, shows the lowest PPL of 6.55 at a $30\%$ pruning ratio, outperforming both ShortGPT and ShortGPT + LoRA. In terms of zero-shot task performance, FlexiGPT achieves the highest scores in ARC-c (38.62\%), PIQA (74.12\%), WinoGrande (66.78\%), and HellaSwag (69.02\%), with an average performance of 62.68\%. This represents a significant improvement over the other methods, demonstrating the effectiveness of our approach. For the $40\%$ pruning ratio, similar trends are observed as FlexiGPT consistently shows superior performance over other methods, achieving the \emph{highest score in every benchmark task}.

Table~\ref{tab:main_results_prune_8b} summarizes the perplexity (PPL) and zero-shot task performance of various pruning methods on the Llama-3 8B model, and Table~\ref{tab:main_results_prune_opt} summarizes the perplexity (PPL) of various pruning methods on the OPT 6.7B and OPT 1.3B models~\cite{zhang2022opt}. These trends also align with those seen in the Llama-2 7B models, further validating the robustness of our method across different model sizes and pruning ratios. We note that Llama-3 8B is much more sensitive to pruning compared to Llama-2 7B, which \emph{underscores the need for post-pruning recovery} such as our weight-sharing and adapters scheme.

\noindent \textbf{Ablation Results} - Table~\ref{tab:ablations_ppl} presents the results of our ablation studies, highlighting the importance of each component in our pruning method. Removing the weight-sharing score, output normalization, or LoRA initialization leads to higher PPL, confirming that each component contributes to the overall effectiveness of our approach.

\input{tables/results_prune_compute}
\noindent \textbf{Analysis - Computation and Throughput} - Table~\ref{tab:main_results_compute} shows the normalized computation costs and throughputs for running our method compared to ShortGPT~\cite{men2024shortgpt} and the unpruned Llama-2 7B model on a single A100 GPU. The unpruned model serves as the baseline with 100\% computation time and throughput. Our method incurs a marginal increase in compute cost compared to the unpruned model but achieves a reduction in the number of stored parameters by approximately 30\%. Although our method is slower than ShortGPT, this is expected, as our approach involves replacing the pruned blocks with weight-sharing techniques.  However, as shown in Table~\ref{tab:main_results_prune}, our method offers significant performance gains over ShortGPT. These gains come at the expense of compute savings but are crucial for on-device applications that cannot tolerate the performance drop associated with methods like ShortGPT. Our method strikes a balance between computational efficiency and high performance, making it suitable for memory-constrained environments where performance is a critical factor.

In order to increase computational efficiency, we implemented a simple self-speculative decoding where the drafting stage uses FlexiGPT without the weight-sharing replacement layers (i.e., the same architecture as ShortGPT), and the verification stage uses the full FlexiGPT model. Importantly, no extra parameters or heads are needed, and our full model performance is retained. We achieved the same outputs as our model with a speedup of 30.11\% compared to our naïve FlexiGPT decoding. We note that the speedup can be improved by combining our self-speculative decoding with other methods such as Medusa~\cite{cai2024medusa}, Jacobi decoding~\cite{santilli2023acceleratingjacobi}, or speculative decoding~\cite{leviathan2023fast} with a smaller, separate model.

%% file: tables/main-prune.tex
\begin{table*}[t]
\centering
\caption{Perplexity (PPL) and zero-shot task performance of compressed Llama-2 7B models. * indicates the model underwent recovery training for 1B tokens after pruning using the SlimPajamas dataset~\cite{cerebras2023slimpajama}. The results for SliceGPT~\cite{ashkboos2024slicegpt} and LLM Surgeon~\cite{van2023llm} are taken from their papers. Two variants of results are given for LLM Surgeon which correspond to pruning with Wikitext-2~\cite{merity2016pointer} and C4~\cite{2019t5}.}
\resizebox{\textwidth}{!}{
\begin{tabular}{c|c|c|c c c c c|c}

\textbf{Method} & \textbf{Ratio} & \textbf{PPL} & \textbf{ARC-e} & \textbf{ARC-c} & \textbf{PIQA} & \textbf{WinoG.} & \textbf{HellaS.} & \textbf{Average} \\ \hline
Unpruned & 0.0\% & 5.11 & 74.58\% & 46.25\% & 79.11\% & 69.14\% & 76.00\% & 69.02\% \\ \hline
SliceGPT & $30\%$ & N/A & 51.77\% & 31.23\% & 63.55\% & 61.33\% & 49.62\% & 51.50\% \\ 
LLM Surgeon (C4) & $30\%$ & N/A & 62.16\% & 34.47\% & 72.85\% & 56.83\% & 58.11\% & 56.88\% \\ 
LLM Surgeon (Wikitext-2) & $30\%$ & N/A & \bf{63.09\%} & 36.69\% & 73.56\% & 61.09\% & 60.72\% & 59.03\% \\ 
ShortGPT & $30\%$ & 22.76 & 48.61\% & 32.68\% & 64.42\% & 64.33\% & 56.15\% & 53.24\% \\ 
ShortGPT + LoRA* & $30\%$ & 6.71 & 62.50\% & 37.54\% & 73.17\% & 66.61\% & 68.19\% & 61.40\% \\ 
\bf{FlexiGPT}* & $30\%$ & \bf{6.55} & 62.84\% & \bf{38.62\%} & \bf{74.12\%} & \bf{66.78\%} & \bf{69.02\%} & \bf{62.68\%} \\ 
\hline
LLM Surgeon (C4) & $40\%$ & N/A & 51.56\% & 27.99\% & 68.93\% & 55.64\% & 48.10\% & 50.44\% \\ 
LLM Surgeon (Wikitext-2) & $40\%$ & N/A & 52.31\% & 30.29\% & 69.26\% & 54.38\% & 48.04\% & 50.86\% \\ 
ShortGPT & $40\%$ & 42.69 & 41.29\% & 30.03\% & 60.17\% & 60.54\% & 43.72\% & 47.15\% \\ 
ShortGPT + LoRA* & $40\%$ & 7.69 & 55.85\% & 33.11\% & 70.51\% & 65.27\% & 62.02\% & 57.35\% \\ 
\bf{FlexiGPT}* & $40\%$ & \bf{7.35} & \bf{57.03}\% & \bf{33.62}\% & \bf{71.44}\%  & \bf{66.61}\% & \bf{63.22}\% & \bf{58.38}\% \\  
\end{tabular}
}
\label{tab:main_results_prune}
\end{table*}

\begin{table*}[t]
\centering
\caption{Perplexity (PPL) and zero-shot task performance of compressed Llama-3 8B models. * indicates the model underwent recovery training for 1B tokens after pruning using the SlimPajamas dataset~\cite{cerebras2023slimpajama}.}
\resizebox{.85\textwidth}{!}{
\begin{tabular}{c|c|c|c c c c c|c}

\textbf{Method} & \textbf{Ratio} & \textbf{PPL} & \textbf{ARC-e} & \textbf{ARC-c} & \textbf{PIQA} & \textbf{WinoG.} & \textbf{HellaS.} & \textbf{Average} \\ \hline
Unpruned & 0.0\% & 6.30 & 77.69\% & 55.33\% & 80.79\% & 72.85\% & 79.17\% & 73.17\% \\ \hline
ShortGPT & $30\%$ & 1.4e4 & 38.80\% & 31.83\% & 60.83\% & 57.93\% & 31.62\% & 44.20\% \\ 
\bf{FlexiGPT}* & $30\%$ & \bf{8.67} & \bf{64.02}\% & \bf{41.21}\% & \bf{74.76}\% & \bf{70.09}\% & \bf{69.12}\% & \bf{63.85}\% \\ 
\hline
ShortGPT & $40\%$ & 9.1e4 & 36.99\% & 30.20\% & 58.60\% & 54.85\% & 30.72\% & 42.27\% \\ 
\bf{FlexiGPT}* & $40\%$ & \bf{10.25} & \bf{55.60}\% & \bf{37.88}\% & \bf{69.31}\% & \bf{66.14}\% & \bf{59.60}\% & \bf{57.70}\% \\ 
\end{tabular}
}
\label{tab:main_results_prune_8b}
\end{table*}

\begin{table}[t]
\centering
\caption{Perplexity (PPL) of compressed OPT models. * indicates the model underwent recovery training for 1B tokens after pruning using the SlimPajamas dataset~\cite{cerebras2023slimpajama}.}
\resizebox{.47\textwidth}{!}{
\begin{tabular}{c|c|c|c}

\textbf{Method} & \textbf{Ratio} & \textbf{OPT 6.7B} & \textbf{OPT 1.3B} \\ \hline
Unpruned & 0.0\% & 7.46  & 9.29 \\ 
\hline
ShortGPT & $30\%$ & 8.61e2 & 6.26e2 \\ 
ShortGPT + FT* & $30\%$ & 8.66 & 11.04 \\ 
\bf{FlexiGPT}* & $30\%$ & \bf{8.39} & \bf{10.81} \\ \hline
ShortGPT & $40\%$ & 2.38e3 & 1.19e3 \\ 
ShortGPT + FT* & $40\%$ & 10.12 & 13.25 \\ 
\bf{FlexiGPT}* & $40\%$ & \bf{9.18} & \bf{11.54} \\ 

\end{tabular}
}
\label{tab:main_results_prune_opt}
\end{table}

\begin{table}[t]
\centering
\caption{Ablation Perplexity (PPL) of $30\%$ compressed Llama-2 7B models. The models underwent recovery training for 1B tokens after pruning using the SlimPajamas dataset~\cite{cerebras2023slimpajama}. We include post-prune PPL (denoted as Start PPL) to show the effect of output feature normalization and adapter initialization on starting PPL.}
\resizebox{.47\textwidth}{!}{
\begin{tabular}{c|c c}
\textbf{Method} & \textbf{Start PPL} & \textbf{PPL} \\ \hline
Ablate high-rank prune (\ref{sec:method:index}) & 22.54 & 6.77 \\ 
Ablate output norm. (\ref{sec:method:norm}) & 8648.94 & 6.68 \\ 
Ablate LoRA init. (\ref{sec:method:adapters}) & 19.69 & 6.63 \\ 
\bf{Full Method} & \bf{21.82} & \bf{6.55} \\
\end{tabular}
}
\label{tab:ablations_ppl}
\end{table}

%% file: tables/main-extend.tex
\begin{table*}[t]
\centering
\caption{Perplexity (PPL) and zero-shot task performance of extended TinyLlama 1.1B models. All models underwent continued pre-training on 10B tokens from the SlimPajamas dataset~\cite{cerebras2023slimpajama}.}
\resizebox{\textwidth}{!}{
\begin{tabular}{c|c|c|c c c c c|c}

\textbf{Method} & \textbf{Layers} & \textbf{PPL} & \textbf{ARC-e} & \textbf{ARC-c} & \textbf{PIQA} & \textbf{WinoG.} & \textbf{HellaS.} & \textbf{Average} \\ \hline
Base & 22 & 6.84 & 55.34\% & 30.11\% & 73.29\% & 59.11\% & 59.20\% & 55.41\% \\ 
\hline
\bf{FlexiGPT (Block)} & 36 & \bf{6.73} & 56.90\% & \bf{31.48\%} & 73.23\% & \bf{59.28\%} & \bf{59.77\%} & \bf{56.13\%} \\ 
\bf{FlexiGPT (Sequential)} & 36 & 6.76 & \bf{56.94\%} & 30.72\% & \bf{73.78\%} & 57.85\% & 59.32\% & 55.72\% \\  
\end{tabular}
}
\label{tab:main_results_extend}
\end{table*}

%% file: tables/results_prune_compute.tex
\begin{table}[t]
\centering
\caption{Normalized computation costs and throughputs for 1xA100 running FlexiGPT vs ShortGPT vs Unpruned on the Llama-2 7B model.}
\resizebox{.4\textwidth}{!}{
\begin{tabular}{c|c|c}
\textbf{Method} & \textbf{Norm. Time} & \textbf{Throughput} \\ \hline
Unpruned & 100.0\% & 100.0\% \\ 
ShortGPT & 65.4\% & 152.8\% \\ 
FlexiGPT & 105.1\% & 95.1\% \\ 
\end{tabular}
}
\label{tab:main_results_compute}
\end{table}

%% file: sections/5_experiments-extend.tex
\section{Model Extension with FlexiGPT}
\label{sec:exp_extend}

\subsection{Setup}

In the previous section, we showed that \method~is a powerful solution for \emph{pruning and recovering} LLMs. In this section, we show that \method~can also be used to extend an off-the-shelf LLM and \emph{introduce performance gains with marginal parameter overhead.} We evaluated our method for model extension using TinyLLaMA~\cite{zhang2024tinyllama} due to its suitability for demonstrating the effectiveness of our approach in extending smaller models. The resources, framework, datasets, and benchmarks are the same as the previous section.

\subsection{Results}

\noindent \textbf{Main Results} - Table~\ref{tab:main_results_extend} shows the perplexity (PPL) and zero-shot task performance of extended TinyLLaMA 1.1B models after continued pre-training on 1B tokens from the SlimPajamas dataset. The base model with 22 layers serves as our baseline.

Our method, FlexiGPT, was evaluated with two extension strategies: Block and Sequential. Both strategies extend the model to 36 layers. FlexiGPT (Block) achieves the lowest PPL of 6.73, compared to the base model's 6.84, indicating a more efficient model. In terms of zero-shot task performance, FlexiGPT (Block) consistently outperforms the base model across most tasks, with notable improvements in ARC-e (56.90\% vs. 55.34\%), ARC-c (31.48\% vs. 30.11\%), and HellaSwag (59.77\% vs. 59.20\%). FlexiGPT (Sequential) also shows competitive results with a PPL of 6.76. It achieves the highest performance in ARC-e (56.94\%) and PIQA (73.78\%) among the extended models. While it slightly underperforms compared to FlexiGPT (Block) in ARC-c and HellaSwag, its overall average performance of 55.72\% still surpasses the baseline. While the downstream task accuracy margins are not as large as the last section, these results are highly significant in that \emph{we are able to boost performance on all tasks using only 10B training tokens for a model which as already been trained on 30T tokens ($\approx$0.3\% extended training).}

\input{tables/results_extend_compute}
\noindent \textbf{Analysis - Computation and Throughput} - Table~\ref{tab:extend_results_compute} compares the normalized computation costs and throughputs for running our method against TinyLlama 1.1B on a single A100 GPU. The base model serves as the baseline with 100\% computation time and throughput. As expected, our method introduces an increased computation cost due to the extended effective length of our model, which is over $50\%$ longer. However, these costs can be mitigated through strategies such as speculative decoding~\cite{leviathan2023fast} or early-exit~\cite{chen2023ee,elhoushi2024layer,pan2024ee}, where the model is only extended when encountering particularly difficult tasks or data, effectively reducing the overall computation burden.

%% file: tables/results_extend_compute.tex
\begin{table}[t]
\centering
\caption{Normalized computation costs and throughputs for 1xA100 running FlexiGPT vs Unpruned on the TinyLlama 1.1B model.}
\resizebox{.4\textwidth}{!}{
\begin{tabular}{c|c|c}
\textbf{Method} & \textbf{Norm. Time} & \textbf{Throughput} \\ \hline
Base & 100.0\% & 100.0\% \\ 
FlexiGPT & 139.1\% & 71.9\% \\ 
\end{tabular}
}
\label{tab:extend_results_compute}
\end{table}

%% file: sections/6_conclusion.tex
\section{Conclusion}
\label{conclusion}

In this paper, we presented an approach to pruning and extending LLMs using LoRA and weight-sharing techniques. Our method targets memory-constrained devices by selectively pruning model blocks based on an importance score and replacing them with a low-parameter replacement strategy. Empirical evaluations show substantial performance gains over existing methods, highlighting our technique's effectiveness. Furthermore, our approach can extend smaller models, achieving significant performance improvements with minimal additional parameters. This work paves the way for more accessible and efficient on-device NLP applications, leveraging our novel combination of pruning, weight-sharing, and parameter-efficient adapters, thereby bringing the power of LLMs to a broader range of memory-constrained devices and use cases.

\section{Limitations}
\label{limitations}

Our method, while effective in achieving parameter efficiency, does not provide gains in computational efficiency. The focus is primarily on reducing the model size for memory-constrained environments, which means that the computational load remains similar to the unpruned model during inference. Additionally, our approach involves a small post-pruning recovery phase where the model undergoes fine-tuning to regain performance. While this phase is crucial for restoring performance, it does require additional computational resources and time. 

Our study was limited to evaluating three popular models, which may not cover the full spectrum of LLM architectures. However, the principles of our method are broadly applicable, and we have no reason to believe the results would not extrapolate to other models with similar architectures. Future work could involve testing our method on a wider variety of models to further validate its generalizability.

\section{Broader Impact}
\label{app:impact}

\looseness=-1
Our method emphasizes parameter efficiency over computation efficiency, making it particularly valuable for on-device settings where memory and storage constraints are critical. By reducing the model size without significantly impacting performance, our approach enables the deployment of powerful LLMs on devices with limited resources, such as smartphones and edge devices. This can democratize access to advanced NLP capabilities, bringing sophisticated language understanding and generation tools to a broader range of users and applications.

Furthermore, our method can be used in conjunction with faster models, deploying the pruned model only for more complex tasks. This hybrid approach can virtually eliminate the computation cost on average while boosting performance for difficult tasks, requiring minimal parameter overhead. This flexibility in deployment can lead to more efficient and effective use of LLMs in various real-world applications.

\section{Potential Risks}
\label{app:risks}

While our work is designed to move LLMs to on-device settings, thereby increasing security and data privacy, there are some potential risks. One risk is that our method involves a small post-training phase, unlike many one-shot pruning methods. This post-training phase could contribute to environmental impact as it requires additional compute, albeit to a smaller extent compared to the initial training of LLMs. Additionally, the ability to deploy LLMs on a wider range of devices could inadvertently lead to increased surveillance. Lastly, while our method emphasizes parameter efficiency, it does not address computational efficiency during inference, which might still pose challenges for extremely resource-constrained environments.

%% file: figures/method_prune_svd-rank-analysis_b.tex
\begin{figure*}[t]
    \centering
    \subfigure[Eq. \eqref{eq:ws-score} \emph{with} high-rank pruning]{
        \includegraphics[width=0.31\textwidth]{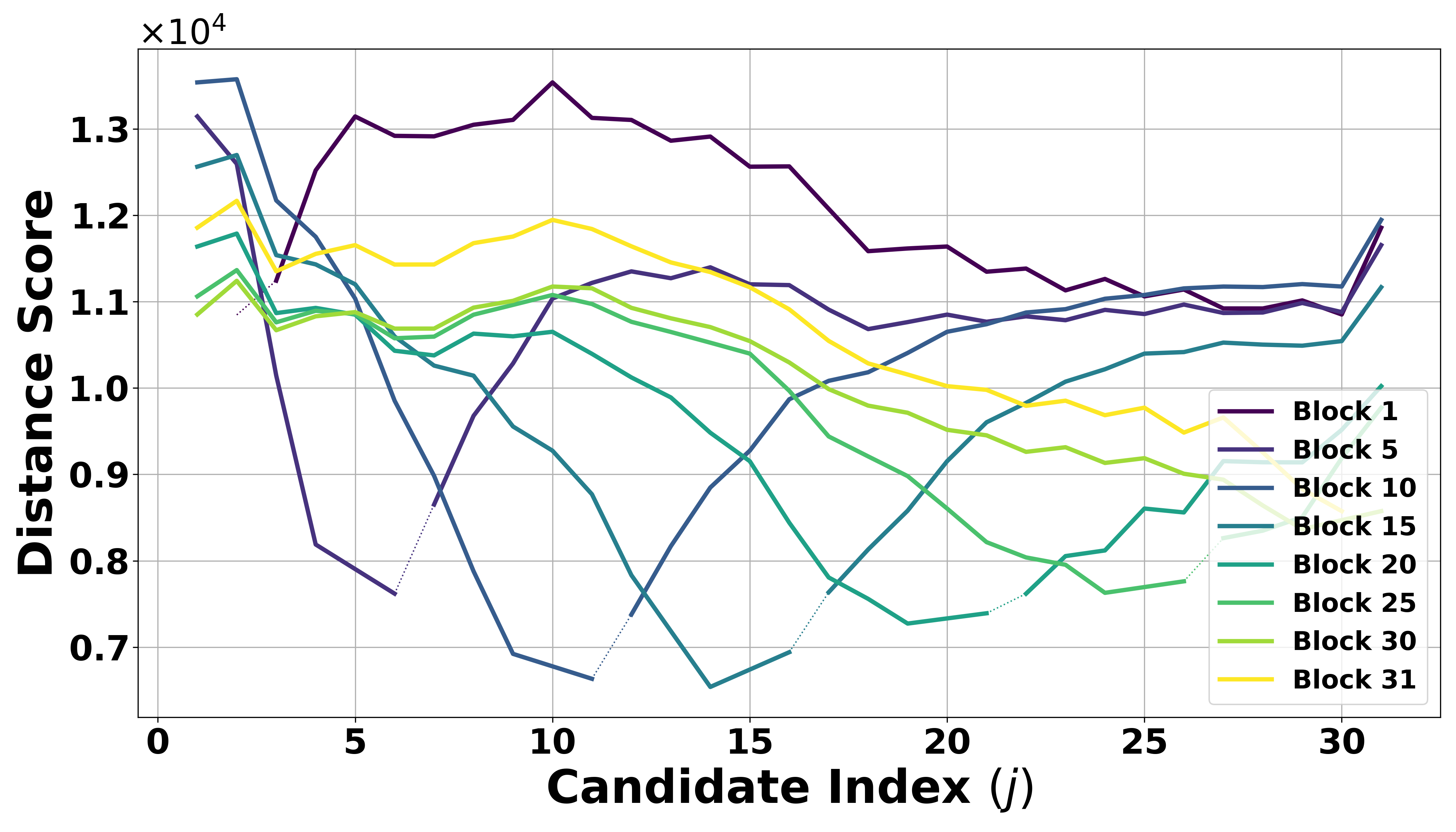}
        \label{fig:prune-svd-rank-analysis-a_b}
    }
    \hfill
    \subfigure[Eq. \eqref{eq:ws-score} \emph{without} high-rank pruning]{
        \includegraphics[width=0.31\textwidth]{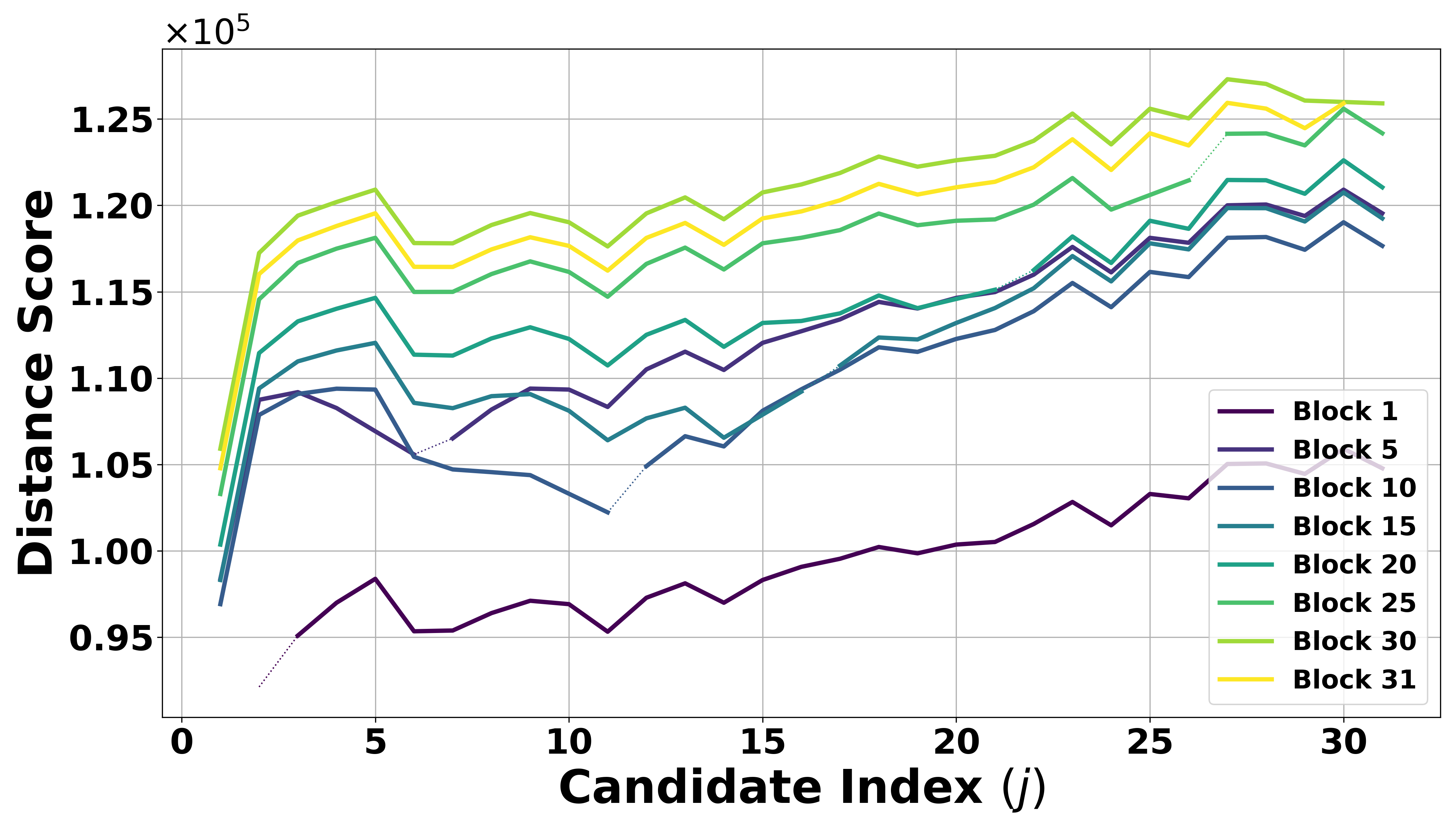}
        \label{fig:prune-svd-rank-analysis-b_b}
    }
    \hfill
    \subfigure[Frobenius norm of $\mathbf{W}_i - \mathbf{W}_j$]{
        \includegraphics[width=0.31\textwidth]{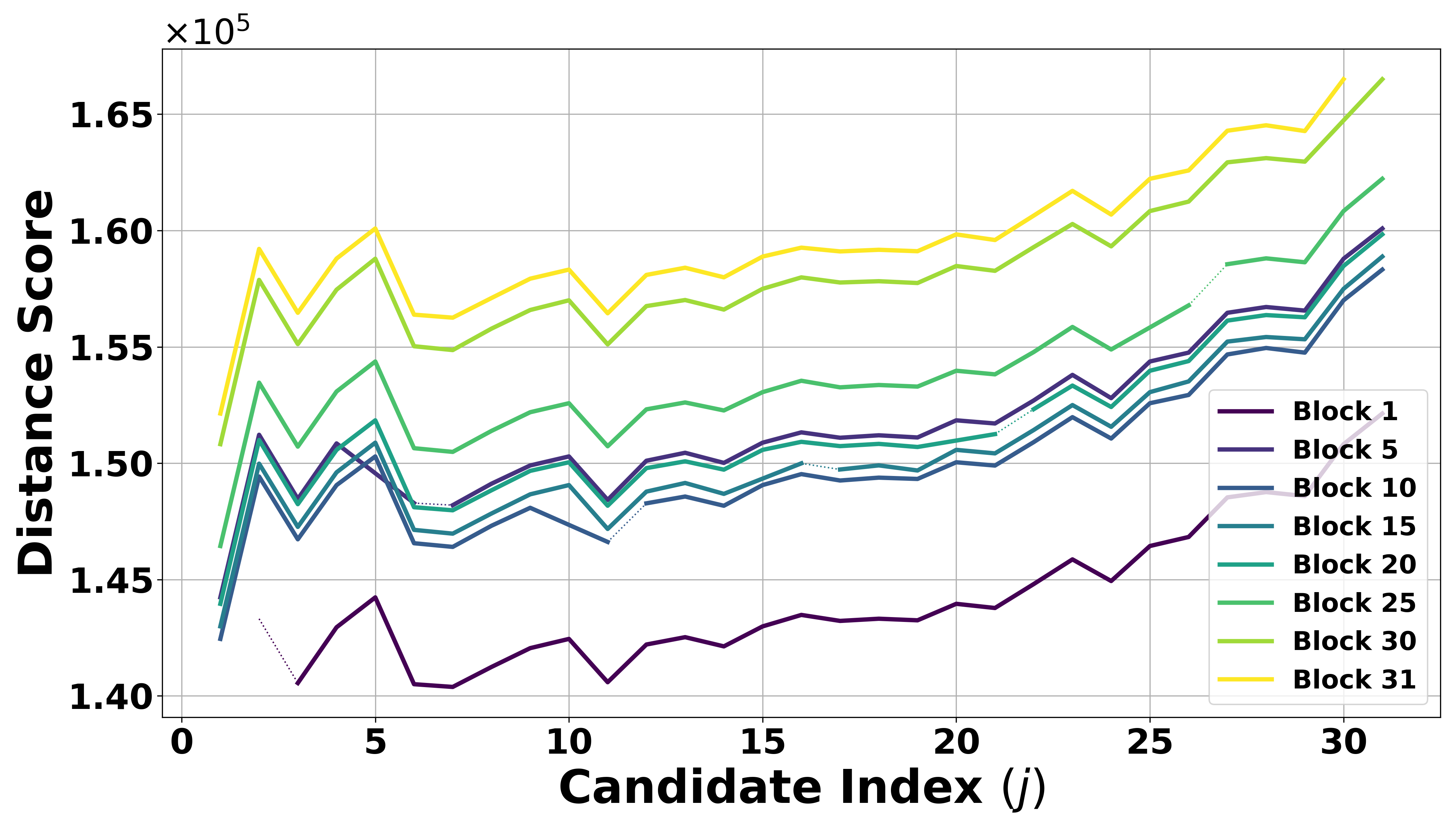}
        \label{fig:prune-svd-rank-analysis-c_b}
    }
    \caption{Comparison of block distance score versus candidate block index $j$ for different metrics. Dotted lines represent where candidate block index $j$ is equal to pruning block index $i$, which is not a valid candidate. (a) Using the proposed metric in Eq. \eqref{eq:ws-score} with high-rank pruning, showing that closer blocks score lower (better), matching our intuition that weights close in the model have similar function. (b) Ablation of high-rank pruning, where there is no clear trend except that blocks closer to 0 are lower and those closer to 31 are higher. (c) Simple Frobenius norm, showing a similar lack of clear trend as in (b). \emph{We found that using the score in (a) as the weight-sharing selection metric results in a much higher performing model compared to using the scores in (b) and (c).}}
    \label{fig:prune-svd-rank-analysis_b}
\end{figure*}

%% file: sections/7_appendix.tex
\section{Additional Experimental Details}
\label{app:exp_details}

Our implementations were carried out using PyTorch, utilizing Fully Sharded Data Parallel (FSDP) and FP-16 mixed precision training for enhanced efficiency. The experiments were conducted on a setup comprising 4 NVIDIA A100 80GB GPUs and required $\approx 192$ gpu hours per experiment. While we only report a single run per result, we evaluate on several models and several tasks. For model handling and training, we employed the Hugging Face Transformers library. We used a learning rate of 0.004 with a a cosine learning rate decay schedule, with a batch size of 2 per GPU and a total batch size of 480 achieved through gradient accumulation. The SlimPajamas dataset~\cite{cerebras2023slimpajama} train set was used, with 1B tokens dedicated to pruning experiments and 10B tokens for model extension experiments due to the faster processing speeds of the models. The LoRA rank utilized was 256. Compared to ShortGPT, our method incurs a 3.67\% relative increase in total parameters for the main experiment setting of Table~\ref{tab:main_results_prune}.

\section{Additional Method Analysis}

In Figure~\ref{fig:prune-svd-rank-analysis_b}, we show an alternative version of Figure~\ref{fig:prune-svd-rank-analysis} where the x-axis is candidate block index $j$ instead of block index distance ($i-j$). The purpose of this figure is to give an additional way to visualize the metrics which better highlights the issue in Figures~\ref{fig:prune-svd-rank-analysis-b} and \ref{fig:prune-svd-rank-analysis-c} where all blocks $i$ choose candidate block $j=0$.

\section{Licenses of Datasets and Models}
\label{app:license}

We used 1B tokens from the SlimPajama~\cite{cerebras2023slimpajama} pre-training dataset for post-prune recovery. For zero-shot performance evaluations, we used the ARC-e, ARC-c~\cite{clark2018think}, PIQA~\cite{bisk2020piqa}, WinoGrande~\cite{sakaguchi2021winogrande}, and HellaSwag~\cite{zellers2019hellaswag} zero-shot benchmarks, utilizing the LM Evaluation Harness~\cite{gao2021framework}. For perplexity performance evaluations, we used the validation MiniPile~\cite{kaddour2023minipile} subset of the Pile dataset~\cite{gao2020pile}. We confirmed that the data that was used does not contain any information that names or uniquely identifies individual people or offensive content by checking their distribution sources. All datasets use the English language. For the pruning experiments, we evaluated our method using LLaMA-2 7B~\cite{touvron2023llama}, OPT 1.3B and 6.7B~\cite{zhang2022opt}, and LLaMA-3 8B~\cite{llama3modelcard}, focusing on these due to their widespread adoption by the community. For the model extension experiments, we evaluated our method for model extension using TinyLLaMA~\cite{zhang2024tinyllama} due to its suitability for demonstrating the effectiveness of our approach in extending smaller models.

The licenses for the datasets and models used in this paper are as follows:
\begin{itemize}
    \item \textbf{SlimPajama}: Apache License 2.0
    \item \textbf{ARC}: CC BY-SA 4.0
    \item \textbf{PIQA}: Academic Free License 3.0
    \item \textbf{HellaSwag}: MIT License
    \item \textbf{WinoGrande}: Apache License 2.0
    \item \textbf{MiniPile}: MIT License
    \item \textbf{LLaMA}: Meta LLaMA Community License Agreement
    \item \textbf{OPT}: OPT License Agreement
    \item \textbf{TinyLLaMA}: Apache License 2.0
\end{itemize}

We used the datasets and models purely for scientific research purposes to create this paper, which is within the scope of their licenses and intended uses.

%% file: main.bbl
\begin{thebibliography}{44}
\providecommand{\natexlab}[1]{#1}

\bibitem[{AI@Meta(2024)}]{llama3modelcard}
AI@Meta. 2024.
\newblock Llama 3 model card.
\newblock \emph{https://github.com/meta-llama/llama3/blob/main}.

\bibitem[{Ashkboos et~al.(2024)Ashkboos, Croci, Nascimento, Hoefler, and Hensman}]{ashkboos2024slicegpt}
Saleh Ashkboos, Maximilian~L Croci, Marcelo Gennari~do Nascimento, Torsten Hoefler, and James Hensman. 2024.
\newblock Slicegpt: Compress large language models by deleting rows and columns.
\newblock \emph{arXiv preprint arXiv:2401.15024}.

\bibitem[{Ba et~al.(2016)Ba, Kiros, and Hinton}]{ba2016layer}
Jimmy~Lei Ba, Jamie~Ryan Kiros, and Geoffrey~E Hinton. 2016.
\newblock Layer normalization.
\newblock \emph{arXiv preprint arXiv:1607.06450}.

\bibitem[{Bisk et~al.(2020)Bisk, Zellers, Gao, Choi et~al.}]{bisk2020piqa}
Yonatan Bisk, Rowan Zellers, Jianfeng Gao, Yejin Choi, et~al. 2020.
\newblock Piqa: Reasoning about physical commonsense in natural language.
\newblock In \emph{Proceedings of the AAAI conference on artificial intelligence}, volume~34, pages 7432--7439.

\bibitem[{Cai et~al.(2024)Cai, Li, Geng, Peng, Lee, Chen, and Dao}]{cai2024medusa}
Tianle Cai, Yuhong Li, Zhengyang Geng, Hongwu Peng, Jason~D Lee, Deming Chen, and Tri Dao. 2024.
\newblock Medusa: Simple llm inference acceleration framework with multiple decoding heads.
\newblock \emph{arXiv preprint arXiv:2401.10774}.

\bibitem[{Cao et~al.(2024)Cao, Yang, and Zhao}]{cao2024head}
Zouying Cao, Yifei Yang, and Hai Zhao. 2024.
\newblock Head-wise shareable attention for large language models.
\newblock \emph{arXiv preprint arXiv:2402.11819}.

\bibitem[{Chen et~al.(2023)Chen, Pan, Li, Ding, and Zhou}]{chen2023ee}
Yanxi Chen, Xuchen Pan, Yaliang Li, Bolin Ding, and Jingren Zhou. 2023.
\newblock Ee-llm: Large-scale training and inference of early-exit large language models with 3d parallelism.
\newblock \emph{arXiv preprint arXiv:2312.04916}.

\bibitem[{Clark et~al.(2018)Clark, Cowhey, Etzioni, Khot, Sabharwal, Schoenick, and Tafjord}]{clark2018think}
Peter Clark, Isaac Cowhey, Oren Etzioni, Tushar Khot, Ashish Sabharwal, Carissa Schoenick, and Oyvind Tafjord. 2018.
\newblock Think you have solved question answering? try arc, the ai2 reasoning challenge.
\newblock \emph{arXiv preprint arXiv:1803.05457}.

\bibitem[{Dehghani et~al.(2018)Dehghani, Gouws, Vinyals, Uszkoreit, and Kaiser}]{dehghani2018universal}
Mostafa Dehghani, Stephan Gouws, Oriol Vinyals, Jakob Uszkoreit, and {\L}ukasz Kaiser. 2018.
\newblock Universal transformers.
\newblock \emph{arXiv preprint arXiv:1807.03819}.

\bibitem[{Denton et~al.(2014)Denton, Zaremba, Bruna, LeCun, and Fergus}]{denton2014exploiting}
Emily~L Denton, Wojciech Zaremba, Joan Bruna, Yann LeCun, and Rob Fergus. 2014.
\newblock Exploiting linear structure within convolutional networks for efficient evaluation.
\newblock \emph{Advances in neural information processing systems}, 27.

\bibitem[{Elhoushi et~al.(2024)Elhoushi, Shrivastava, Liskovich, Hosmer, Wasti, Lai, Mahmoud, Acun, Agarwal, Roman et~al.}]{elhoushi2024layer}
Mostafa Elhoushi, Akshat Shrivastava, Diana Liskovich, Basil Hosmer, Bram Wasti, Liangzhen Lai, Anas Mahmoud, Bilge Acun, Saurabh Agarwal, Ahmed Roman, et~al. 2024.
\newblock Layer skip: Enabling early exit inference and self-speculative decoding.
\newblock \emph{arXiv preprint arXiv:2404.16710}.

\bibitem[{Gao et~al.(2020)Gao, Biderman, Black, Golding, Hoppe, Foster, Phang, He, Thite, Nabeshima et~al.}]{gao2020pile}
Leo Gao, Stella Biderman, Sid Black, Laurence Golding, Travis Hoppe, Charles Foster, Jason Phang, Horace He, Anish Thite, Noa Nabeshima, et~al. 2020.
\newblock The {P}ile: An 800{GB} dataset of diverse text for language modeling.
\newblock \emph{arXiv preprint arXiv:2101.00027}.

\bibitem[{Gao et~al.(2021)Gao, Tow, Biderman, Black, DiPofi, Foster, Golding, Hsu, McDonell, Muennighoff et~al.}]{gao2021framework}
Leo Gao, Jonathan Tow, Stella Biderman, Sid Black, Anthony DiPofi, Charles Foster, Laurence Golding, Jeffrey Hsu, Kyle McDonell, Niklas Muennighoff, et~al. 2021.
\newblock A framework for few-shot language model evaluation.
\newblock \emph{Version v0. 0.1. Sept}, page~8.

\bibitem[{Hadi et~al.(2023)Hadi, Qureshi, Shah, Irfan, Zafar, Shaikh, Akhtar, Wu, Mirjalili et~al.}]{hadi2023large}
Muhammad~Usman Hadi, Rizwan Qureshi, Abbas Shah, Muhammad Irfan, Anas Zafar, Muhammad~Bilal Shaikh, Naveed Akhtar, Jia Wu, Seyedali Mirjalili, et~al. 2023.
\newblock Large language models: a comprehensive survey of its applications, challenges, limitations, and future prospects.
\newblock \emph{Authorea Preprints}.

\bibitem[{He et~al.(2021)He, Zhou, Ma, Berg-Kirkpatrick, and Neubig}]{he2021towards}
Junxian He, Chunting Zhou, Xuezhe Ma, Taylor Berg-Kirkpatrick, and Graham Neubig. 2021.
\newblock Towards a unified view of parameter-efficient transfer learning.
\newblock \emph{arXiv preprint arXiv:2110.04366}.

\bibitem[{Houlsby et~al.(2019)Houlsby, Giurgiu, Jastrzebski, Morrone, De~Laroussilhe, Gesmundo, Attariyan, and Gelly}]{houlsby2019parameter}
Neil Houlsby, Andrei Giurgiu, Stanislaw Jastrzebski, Bruna Morrone, Quentin De~Laroussilhe, Andrea Gesmundo, Mona Attariyan, and Sylvain Gelly. 2019.
\newblock Parameter-efficient transfer learning for nlp.
\newblock In \emph{International Conference on Machine Learning}, pages 2790--2799. PMLR.

\bibitem[{Hsu et~al.(2022)Hsu, Hua, Chang, Lou, Shen, and Jin}]{hsu2022language}
Yen-Chang Hsu, Ting Hua, Sungen Chang, Qian Lou, Yilin Shen, and Hongxia Jin. 2022.
\newblock Language model compression with weighted low-rank factorization.
\newblock \emph{arXiv preprint arXiv:2207.00112}.

\bibitem[{Hu et~al.(2021)Hu, Shen, Wallis, Allen-Zhu, Li, Wang, Wang, and Chen}]{hu2021lora}
Edward~J Hu, Yelong Shen, Phillip Wallis, Zeyuan Allen-Zhu, Yuanzhi Li, Shean Wang, Lu~Wang, and Weizhu Chen. 2021.
\newblock Lora: Low-rank adaptation of large language models.
\newblock \emph{arXiv preprint arXiv:2106.09685}.

\bibitem[{Kaddour(2023)}]{kaddour2023minipile}
Jean Kaddour. 2023.
\newblock The minipile challenge for data-efficient language models.
\newblock \emph{arXiv preprint arXiv:2304.08442}.

\bibitem[{Lester et~al.(2021)Lester, Al-Rfou, and Constant}]{lester2021power}
Brian Lester, Rami Al-Rfou, and Noah Constant. 2021.
\newblock The power of scale for parameter-efficient prompt tuning.
\newblock \emph{arXiv preprint arXiv:2104.08691}.

\bibitem[{Leviathan et~al.(2023)Leviathan, Kalman, and Matias}]{leviathan2023fast}
Yaniv Leviathan, Matan Kalman, and Yossi Matias. 2023.
\newblock Fast inference from transformers via speculative decoding.
\newblock In \emph{International Conference on Machine Learning}, pages 19274--19286. PMLR.

\bibitem[{Liu et~al.(2024{\natexlab{a}})Liu, Wang, Yin, Molchanov, Wang, Cheng, and Chen}]{liu2024dora}
Shih-Yang Liu, Chien-Yi Wang, Hongxu Yin, Pavlo Molchanov, Yu-Chiang~Frank Wang, Kwang-Ting Cheng, and Min-Hung Chen. 2024{\natexlab{a}}.
\newblock Dora: Weight-decomposed low-rank adaptation.
\newblock \emph{arXiv preprint arXiv:2402.09353}.

\bibitem[{Liu et~al.(2021)Liu, Ji, Fu, Tam, Du, Yang, and Tang}]{liu2021p}
Xiao Liu, Kaixuan Ji, Yicheng Fu, Weng~Lam Tam, Zhengxiao Du, Zhilin Yang, and Jie Tang. 2021.
\newblock P-tuning v2: Prompt tuning can be comparable to fine-tuning universally across scales and tasks.
\newblock \emph{arXiv preprint arXiv:2110.07602}.

\bibitem[{Liu et~al.(2024{\natexlab{b}})Liu, Zhao, Iandola, Lai, Tian, Fedorov, Xiong, Chang, Shi, Krishnamoorthi et~al.}]{liu2024mobilellm}
Zechun Liu, Changsheng Zhao, Forrest Iandola, Chen Lai, Yuandong Tian, Igor Fedorov, Yunyang Xiong, Ernie Chang, Yangyang Shi, Raghuraman Krishnamoorthi, et~al. 2024{\natexlab{b}}.
\newblock Mobilellm: Optimizing sub-billion parameter language models for on-device use cases.
\newblock \emph{arXiv preprint arXiv:2402.14905}.

\bibitem[{Ma et~al.(2023)Ma, Fang, and Wang}]{ma2023llm}
Xinyin Ma, Gongfan Fang, and Xinchao Wang. 2023.
\newblock Llm-pruner: On the structural pruning of large language models.
\newblock \emph{Advances in neural information processing systems}, 36:21702--21720.

\bibitem[{Men et~al.(2024)Men, Xu, Zhang, Wang, Lin, Lu, Han, and Chen}]{men2024shortgpt}
Xin Men, Mingyu Xu, Qingyu Zhang, Bingning Wang, Hongyu Lin, Yaojie Lu, Xianpei Han, and Weipeng Chen. 2024.
\newblock Shortgpt: Layers in large language models are more redundant than you expect.
\newblock \emph{arXiv preprint arXiv:2403.03853}.

\bibitem[{Merity et~al.(2016)Merity, Xiong, Bradbury, and Socher}]{merity2016pointer}
Stephen Merity, Caiming Xiong, James Bradbury, and Richard Socher. 2016.
\newblock \href {https://arxiv.org/abs/1609.07843} {Pointer sentinel mixture models}.
\newblock \emph{Preprint}, arXiv:1609.07843.

\bibitem[{Minaee et~al.(2024)Minaee, Mikolov, Nikzad, Chenaghlu, Socher, Amatriain, and Gao}]{minaee2024large}
Shervin Minaee, Tomas Mikolov, Narjes Nikzad, Meysam Chenaghlu, Richard Socher, Xavier Amatriain, and Jianfeng Gao. 2024.
\newblock Large language models: A survey.
\newblock \emph{arXiv preprint arXiv:2402.06196}.

\bibitem[{Naveed et~al.(2023)Naveed, Khan, Qiu, Saqib, Anwar, Usman, Barnes, and Mian}]{naveed2023comprehensive}
Humza Naveed, Asad~Ullah Khan, Shi Qiu, Muhammad Saqib, Saeed Anwar, Muhammad Usman, Nick Barnes, and Ajmal Mian. 2023.
\newblock A comprehensive overview of large language models.
\newblock \emph{arXiv preprint arXiv:2307.06435}.

\bibitem[{Pan et~al.(2024)Pan, Chen, Li, Ding, and Zhou}]{pan2024ee}
Xuchen Pan, Yanxi Chen, Yaliang Li, Bolin Ding, and Jingren Zhou. 2024.
\newblock Ee-tuning: An economical yet scalable solution for tuning early-exit large language models.
\newblock \emph{arXiv preprint arXiv:2402.00518}.

\bibitem[{Raffel et~al.(2019)Raffel, Shazeer, Roberts, Lee, Narang, Matena, Zhou, Li, and Liu}]{2019t5}
Colin Raffel, Noam Shazeer, Adam Roberts, Katherine Lee, Sharan Narang, Michael Matena, Yanqi Zhou, Wei Li, and Peter~J. Liu. 2019.
\newblock \href {https://arxiv.org/abs/1910.10683} {Exploring the limits of transfer learning with a unified text-to-text transformer}.
\newblock \emph{arXiv e-prints}.

\bibitem[{Raiaan et~al.(2024)Raiaan, Mukta, Fatema, Fahad, Sakib, Mim, Ahmad, Ali, and Azam}]{raiaan2024review}
Mohaimenul Azam~Khan Raiaan, Md~Saddam~Hossain Mukta, Kaniz Fatema, Nur~Mohammad Fahad, Sadman Sakib, Most Marufatul~Jannat Mim, Jubaer Ahmad, Mohammed~Eunus Ali, and Sami Azam. 2024.
\newblock A review on large language models: Architectures, applications, taxonomies, open issues and challenges.
\newblock \emph{IEEE Access}.

\bibitem[{Reid et~al.(2021)Reid, Marrese-Taylor, and Matsuo}]{reid2021subformer}
Machel Reid, Edison Marrese-Taylor, and Yutaka Matsuo. 2021.
\newblock Subformer: Exploring weight sharing for parameter efficiency in generative transformers.
\newblock \emph{arXiv preprint arXiv:2101.00234}.

\bibitem[{Sakaguchi et~al.(2021)Sakaguchi, Bras, Bhagavatula, and Choi}]{sakaguchi2021winogrande}
Keisuke Sakaguchi, Ronan~Le Bras, Chandra Bhagavatula, and Yejin Choi. 2021.
\newblock Winogrande: An adversarial winograd schema challenge at scale.
\newblock \emph{Communications of the ACM}, 64(9):99--106.

\bibitem[{Santilli et~al.(2023)Santilli, Severino, Postolache, Maiorca, Mancusi, Marin, and Rodol{\`a}}]{santilli2023acceleratingjacobi}
Andrea Santilli, Silvio Severino, Emilian Postolache, Valentino Maiorca, Michele Mancusi, Riccardo Marin, and Emanuele Rodol{\`a}. 2023.
\newblock Accelerating transformer inference for translation via parallel decoding.
\newblock \emph{arXiv preprint arXiv:2305.10427}.

\bibitem[{Soboleva et~al.(2023)Soboleva, Al-Khateeb, Myers, Steeves, Hestness, and Dey}]{cerebras2023slimpajama}
Daria Soboleva, Faisal Al-Khateeb, Robert Myers, Jacob~R Steeves, Joel Hestness, and Nolan Dey. 2023.
\newblock \href {https://huggingface.co/datasets/cerebras/SlimPajama-627B} {Slimpajama: A 627b token cleaned and deduplicated version of redpajama}.
\newblock \url{https://www.cerebras.net/blog/slimpajama-a-627b-token-cleaned-and-deduplicated-version-of-redpajama}.

\bibitem[{Takase and Kiyono(2021)}]{takase2021lessons}
Sho Takase and Shun Kiyono. 2021.
\newblock Lessons on parameter sharing across layers in transformers.
\newblock \emph{arXiv preprint arXiv:2104.06022}.

\bibitem[{Touvron et~al.(2023)Touvron, Lavril, Izacard, Martinet, Lachaux, Lacroix, Rozi{\`e}re, Goyal, Hambro, Azhar et~al.}]{touvron2023llama}
Hugo Touvron, Thibaut Lavril, Gautier Izacard, Xavier Martinet, Marie-Anne Lachaux, Timoth{\'e}e Lacroix, Baptiste Rozi{\`e}re, Naman Goyal, Eric Hambro, Faisal Azhar, et~al. 2023.
\newblock Llama: Open and efficient foundation language models.
\newblock \emph{arXiv preprint arXiv:2302.13971}.

\bibitem[{van~der Ouderaa et~al.(2024)van~der Ouderaa, Nagel, Van~Baalen, and Blankevoort}]{van2023llm}
Tycho~FA van~der Ouderaa, Markus Nagel, Mart Van~Baalen, and Tijmen Blankevoort. 2024.
\newblock The llm surgeon.
\newblock In \emph{The Twelfth International Conference on Learning Representations}.

\bibitem[{Vaswani et~al.(2017)Vaswani, Shazeer, Parmar, Uszkoreit, Jones, Gomez, Kaiser, and Polosukhin}]{vaswani2017attention}
Ashish Vaswani, Noam Shazeer, Niki Parmar, Jakob Uszkoreit, Llion Jones, Aidan~N Gomez, {\L}ukasz Kaiser, and Illia Polosukhin. 2017.
\newblock Attention is all you need.
\newblock \emph{Advances in neural information processing systems}, 30.

\bibitem[{Yang et~al.(2024)Yang, Cao, and Zhao}]{yang2024laco}
Yifei Yang, Zouying Cao, and Hai Zhao. 2024.
\newblock Laco: Large language model pruning via layer collapse.
\newblock \emph{arXiv preprint arXiv:2402.11187}.

\bibitem[{Zellers et~al.(2019)Zellers, Holtzman, Bisk, Farhadi, and Choi}]{zellers2019hellaswag}
Rowan Zellers, Ari Holtzman, Yonatan Bisk, Ali Farhadi, and Yejin Choi. 2019.
\newblock Hellaswag: Can a machine really finish your sentence?
\newblock \emph{arXiv preprint arXiv:1905.07830}.

\bibitem[{Zhang et~al.(2024)Zhang, Zeng, Wang, and Lu}]{zhang2024tinyllama}
Peiyuan Zhang, Guangtao Zeng, Tianduo Wang, and Wei Lu. 2024.
\newblock Tinyllama: An open-source small language model.
\newblock \emph{arXiv preprint arXiv:2401.02385}.

\bibitem[{Zhang et~al.(2022)Zhang, Roller, Goyal, Artetxe, Chen, Chen, Dewan, Diab, Li, Lin et~al.}]{zhang2022opt}
Susan Zhang, Stephen Roller, Naman Goyal, Mikel Artetxe, Moya Chen, Shuohui Chen, Christopher Dewan, Mona Diab, Xian Li, Xi~Victoria Lin, et~al. 2022.
\newblock Opt: Open pre-trained transformer language models.
\newblock \emph{arXiv preprint arXiv:2205.01068}.

\end{thebibliography}
